\documentclass[nopreprintline,12pt]{elsarticle}

\usepackage{fullpage}
\usepackage{subcaption}
\usepackage{graphicx}
\usepackage[table,xcdraw]{xcolor}
\usepackage{amssymb}
\usepackage{amsmath}
\usepackage{amsfonts}
\usepackage{amsthm}
\usepackage{float}
\usepackage{parskip}
\usepackage[english]{babel}
\usepackage[utf8]{inputenc}
\usepackage{hyperref}
\usepackage{fancyhdr}
\usepackage{commath}
\usepackage{mathtools}
\usepackage{multirow}
\usepackage{enumerate}
\usepackage{multicol}
\usepackage{alltt}
\usepackage{relsize}
\usepackage{url}
\usepackage{soul}

\newcommand{\R}{\mathbb{R}}

\def\vec   #1{\mbox{\boldmath $#1$}{}}

\journal{ }

\begin{document}

\begin{frontmatter}



\title{Image Velocimetry using Direct Displacement Field estimation with Neural Networks for fluids}


\author[label1]{Efraín Magaña}
\ead{emmagana@uc.cl}
\author[label1,label2]{Francisco Sahli}
\author[label3]{Wernher Brevis} 

\affiliation[label1]{organization={Department of Mechanical and Metallurgical Engineering,
School of Engineering, Pontificia Universidad Católica de Chile},
            city={Santiago},
            country={Chile}}
\affiliation[label2]{organization={Institute for Biological and Medical Engineering, Schools of Engineering, Medicine and Biological Sciences, Pontificia Universidad Católica de Chile},
            city={Santiago},
            country={Chile}}
\affiliation[label3]{organization={Department of Hydraulic and Environmental Engineering,
School of Engineering, Pontificia Universidad Católica de Chile},
            city={Santiago},
            country={Chile}}

\begin{abstract}
An important tool for experimental fluids mechanics research is Particle Image Velocimetry (PIV). Several robust methodologies have been proposed to perform the estimation of velocity field from the images, however, alternative methods are still needed to increase the spatial resolution of the results. This work presents a novel approach for estimating fluid flow fields using neural networks and the optical flow equation to predict displacement vectors between sequential images. The result is a continuous representation of the displacement, that can be evaluated on the full spatial resolution of the image. The methodology was validated on synthetic and experimental images. Accurate results were obtained in terms of the estimation of instantaneous velocity fields, and of the determined time average turbulence quantities and power spectral density. The methodology proposed differs of previous attempts of using machine learning for this task: it does not require any previous training, and could be directly used in any pair of images. 
\end{abstract}




\end{frontmatter}


\section{Introduction}\label{intro}
Particle Image Velocimetry (PIV) is an essential tool for experimental fluids mechanics research. The technique is used to estimate velocity fields of a flow from a sequence of images of the movement of tracer particles \cite{Raffel2018}. It is undeniable that important contributions have been made by the current PIV technique to the understanding of fluid motion. However, there are still opportunities for improvements, such as the enhancement of the spatial resolution of the calculated velocity field. In PIV, each vector is calculated from the average displacement of particles that move within a pre-defined part of the image called the interrogation window. For example, if a 4096 $\times$ 2160 px (4K resolution) image is considered, the vector field resulting from the use of a uniform 16 $\times$ 16 px interrogation window, will have a size of 256$\times$ 135. Several image processing alternatives have been proposed to improve the resolution of image velocimetry, such as combining PIV with Particle Tracking Velocimetry (PTV) or using other techniques such as optical flow, Kalman filtering \cite{Takehara2000}, or a combination of wavelets transforms and optical flow \cite{Schmidt2019}. However, a velocity field with the same resolution as the image still requires some type of interpolation. 

In the last ten years, multiple attempts have been made to use deep learning models to obtain a velocity field with the same resolution as the input image \cite{cai2019, mingrui2020, Lagemann2021, Changdong2021, Zhang2023, Manickathan2023}, where the approaches have been mainly based on architectures that use layers of convolutional neural networks (CNNs). A CNN layer is a trainable filter that is applied to the image through a convolution. Relevant features in the image are detected by these filters. The application of multiple layers of CNNs results in a hidden state that summarizes the key features of the problem. From the hidden state, a discrete velocity field is reconstructed with the same resolution as the original images. The reconstruction step itself is also trained. Despite the good performance of these architectures, a limitation is present: the resolution of the images must be fixed or highly restricted. In the case of a fixed resolution, this is the product of fixing the size of the hidden state for the reconstruction of the field. As such, when CNN layers are applied to an image without the correct shape, a hidden state with the incorrect dimension will result, which cannot use the learned reconstruction process. Even when the size of the hidden state is not fixed, the trained filters would only be valid for displacements that are similar in magnitude to those used in training, as they are limited to the size of the filter used. 

The size limitation can be addressed, for instance, by slicing the image into multiple images with the desired resolution and then processing them with the trained model. For example, if the image is of 4096 $\times$ 2160 px, assuming no optical distortions, thus a constant magnification between image and real-world coordinates, and assuming that the area captured by the image is 40 $\times$ 21 cm, a slice of 256 $\times$ 256 px (standard resolution of the proposed models) will only represent an area of 2.5 $\times$ 2.5 cm.  If two sequential images are captured with a time interval  $\Delta t = 1 s$, the maximum velocity vector, in the ideal case, cannot be greater than 3.53 [cm/s] (the length of the diagonal of the images). A solution to this limitation was proposed by \cite{Choi2023}, where the images were first deformed using the velocity fields obtained from PIV, causing the new displacements to be smaller, and then a pre-trained CNN was used to obtain the velocity at pixel level. The resulting velocity field corresponded to the combination of the PIV and the CNN estimates, however, the optimal method for aggregating each patch of the resulting displacements is still an open problem. 
Another issue that arises in these models is the necessity for them to be trained with prior examples of correct velocity fields, which creates an implicit bias towards the training data used \cite{Geman1992}. Furthermore, in most cases, the training data used consisted of synthetic PIV images with known velocity fields (except for \cite{mingrui2020}, where unknown velocity fields were used). This means that for a real case, the models can only be trusted if the synthetic data used for training closely resembles the experimental conditions.

As an alternative to current approaches, a simpler methodology is proposed in this work, based on fully-connected neural networks, to learn the velocity fields between a pair of images using the optical flow formulation given by \cite{Berthold1981}, and using only a pair of images without training on prior examples. The use of the optical flow equation and the techniques that derive from it have been previously employed to refine PIV results \cite{Seong2019} and to directly estimate velocity fields with a performance similar to PIV \cite{Mendes2021}. However, an approach is proposed in this work that can be seen as a simpler free mesh variations method as to the proposed in \cite{Ruhnau2005}. Furthermore, the proposed methodology uses a spatial embedding of the image coordinates as the input for the neural network. This allows arbitrary image sizes to be worked and enables the introduction of super-resolution, as the velocity can be estimated in inter-pixel coordinates within the neural network, without the need for further interpolation. 

In a broad sense, the changes of an image into another are easily understood in terms of a displacement field ($\vec{\delta}$) rather than in terms of velocity fields ($\vec{U}$). As a result, the following sections of the work will be based on the analysis of displacements, since the velocity fields can be obtained from ($\vec{\delta}$) after a standard calibration procedure \cite{Quenot2001}. For simplicity, assuming a constant magnification constant between the camera and real-world coordinates $C$ (normally $C$ depends on the coordinates), the velocity field can be defined as:

\begin{equation}
    \vec{U}(x,y) = \frac{C}{\Delta t}\vec{\delta}(x,y)
    \label{eq:conv}
\end{equation}

Where $\Delta t$ is the time between two sequential images.

\section{Methods}\label{meth}
\subsection{Background}
Neural networks are non-linear parametric functions, with parameters $\vec{\theta}$. This function can be decomposed into what is usually referred to as \textit{neurons} and \textit{layers}. Each layer is composed of a number of neurons, and in general, the output of a layer is passed to the next layer. 

In the case of a fully-connected neural network, the neuron is defined as:
\begin{equation}
    y = f\left(\sum_{i=1}^N x_i \theta_i + \theta_0\right)
\end{equation}
Where $f$ is the \textit{activation function} and corresponds to a non-linear function. The variable $x$ corresponds to the input of the neuron, with size $N$, and $y$ is the output. A layer can then be seen as:

\begin{equation}
    \vec{x}_{i+1} = f( \vec{x}_{i} \cdot \vec{\theta}_1^i + \vec{\theta}_0^i) \label{eq:fully}
\end{equation}

Where $\vec{x}_i \in \R^{N_i}$ (a vector), $\vec{\theta}_1^i \in \R^{N_i \times N_{i+1}}$ (a matrix) and $\vec{\theta}_0^i \in \R^{N_{i+1}}$ (a vector), with $N_i$ being the size of the input for the layer ${i+1}$. And $i \in [0, N_l]$ the number of the layer. The output of the final layer is the output of the neural network. For fully-connected neural networks, the parameters represent a weighted sum of the input. In contrast, the dot product on Eq. \ref{eq:fully} is replaced by the convolutional operator ($\ast$) in a CNN, with the parameters representing filters that are applied to the input \cite{wu2017cnn}.

This simple definition allows the parameters $\vec{\theta}$ to be optimised to minimise an objective function, commonly called the \textit{loss function}.

\subsection{Architecture}
The central idea of this methodology is that the vector displacement field $\vec{\delta}$, with components $\delta_x$ on $\hat x$ and $\delta_y$ on $\hat y$, can be approximated by a fully-connected neural network ($\vec{\tilde \delta}$) with parameters $\vec{\theta}$:
\begin{equation}
    \vec{\delta}(x, y)=[\delta_x(x, y), \delta_y(x, y)]\approx \mathit{NN}(x,y;\vec{\theta}) = \vec{ \tilde \delta} (x,y)
\end{equation}
In this form, the inputs of $\vec{\tilde \delta}$ are the image coordinates, and the output is the displacement in each direction. Since the displacement of a fluid can be composed of various frequencies, learning the displacements can be challenging for neural networks since they tend to learn only the dominant frequencies \cite{Rahaman2018}. To avoid this problem, Fourier Features are used in this work as positional embedding for the image coordinates \cite{Tancik2020}. Fourier features take the spatial vector ($\vec{v}=[x,y]$) and embed it with a random Gaussian sampled matrix $\vec{B}$ in a space of sines and cosines with dimensions of $2N_e$, as shown in Eq. \ref{eq:ff}. Thus, $\vec{B} \in M^{N_e\times2}$ and $\vec{B} \sim N(0, {1}/{\beta^2})$, where $\beta$ is used to control the predominant frequencies of the embedding, and in turn those of $\vec{\tilde \delta}$.

\begin{equation}
    \vec{\gamma}(\vec{v})= [\sin(\vec{B}\vec{v}), \cos(\vec{B}\vec{v})]^T
    \label{eq:ff}
\end{equation}

\begin{figure}[H]
    \centering
\begin{subfigure}{\textwidth}
    \centering
    \includegraphics[width=\textwidth]{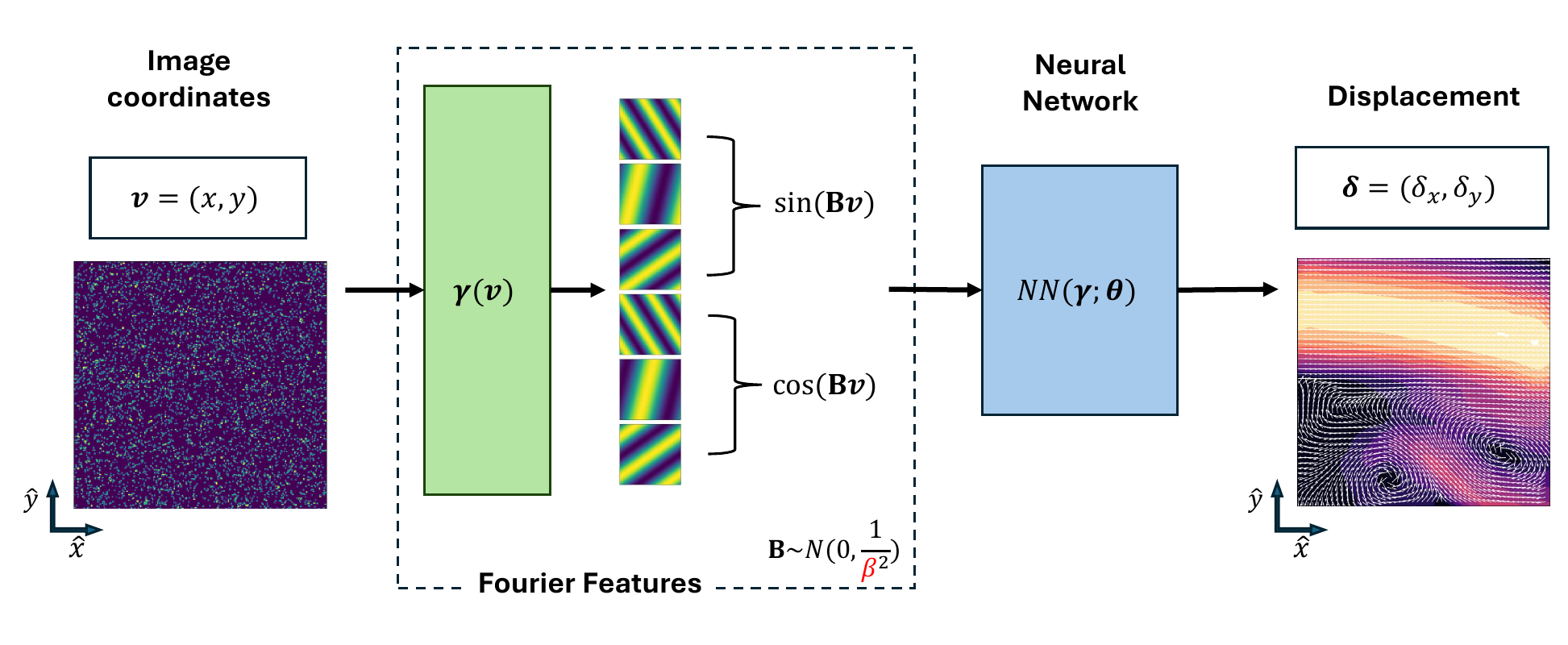}
    \caption{}
    \label{fig:non}
\end{subfigure}
\begin{subfigure}{\textwidth}
    \centering
    \includegraphics[width=.6\textwidth]{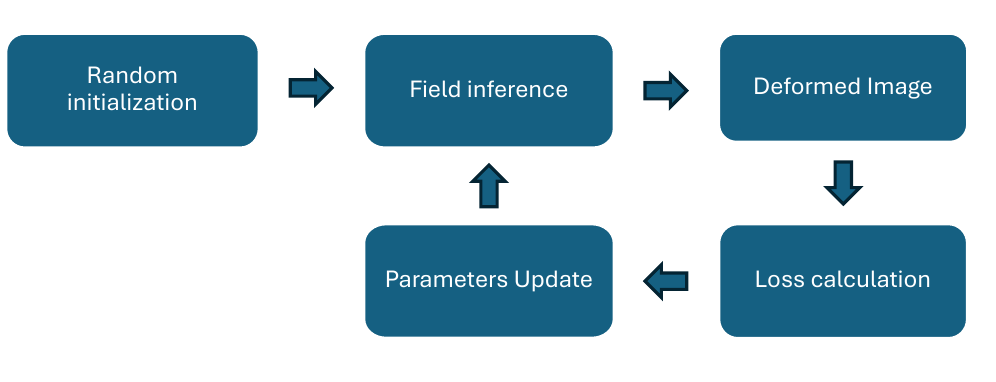}
    \caption{}
    \label{fig:iter}
\end{subfigure}
    \caption{(a) Diagram of the architecture used and (b) iterative learning diagram.}
    \label{fig:diagrams}
\end{figure}

Following the embedding, a fully-connected neural network is used, where the number of layers ($N_l$) and the size of each layer ($S_l$) are defined by the user, as the input and output of the neural network are fixed to $2N_e$ and $2$ respectively. The activation function corresponds to $\tanh(\cdot)$. A schematic diagram of the architecture is presented in Fig. \ref{fig:non}.

\subsection{Unsupervised learning of displacement}
The proposed architecture does not take into account some prior available images. This means that it is not possible for $\vec{\tilde \delta}$ to be trained beforehand with some known data, and thus the displacement vectors are learned in an unsupervised way.

The approach used in this work is based on two sequential images. The second image ($I_2$) is the result of the deformation of the first image ($I_1$), and the deformation is associated with the displacement fields. Then the intensity on $\vec{v}_2=\vec{v}_1+\vec{\delta}(\vec{v}_1)$ in the second image correspond to the intensity on $\vec{v}_1$ in the first image, as shown in the following equation:

\begin{equation}
    I_2(\vec{v}_2)=I_2(\vec{v}_1 + \vec{\delta}(\vec{v}_1))=I_1(\vec{v}_1)
\end{equation}
This assumption forms the core of traditional optical flow \cite{Berthold1981}. Following this, the deformed image con be defined by $\vec{\tilde \delta}$:

\begin{equation}
    I^d(\vec{v}) = I_2(\vec{v}+\vec{\tilde \delta}(\vec{v}))
\end{equation}

Since $I_2$ is an image, and therefore a discrete field, the estimation of $I^d$ is based on the interpolation of the discrete field. In this work, a simple bilinear interpolation is used for this purpose. Following this, the loss function, which is to be minimised optimising $\vec{\theta}$, can be defined as:

\begin{equation}
    \mathcal{L}:=\frac{1}{N_p}\sum_{i=1}^{N_p} (I_1(\vec{v}_i)-I^d(\vec{v}_i))^2
    \label{eq:loss}
\end{equation}

\begin{figure}[H]
    \centering
    \includegraphics[width=.8\textwidth]{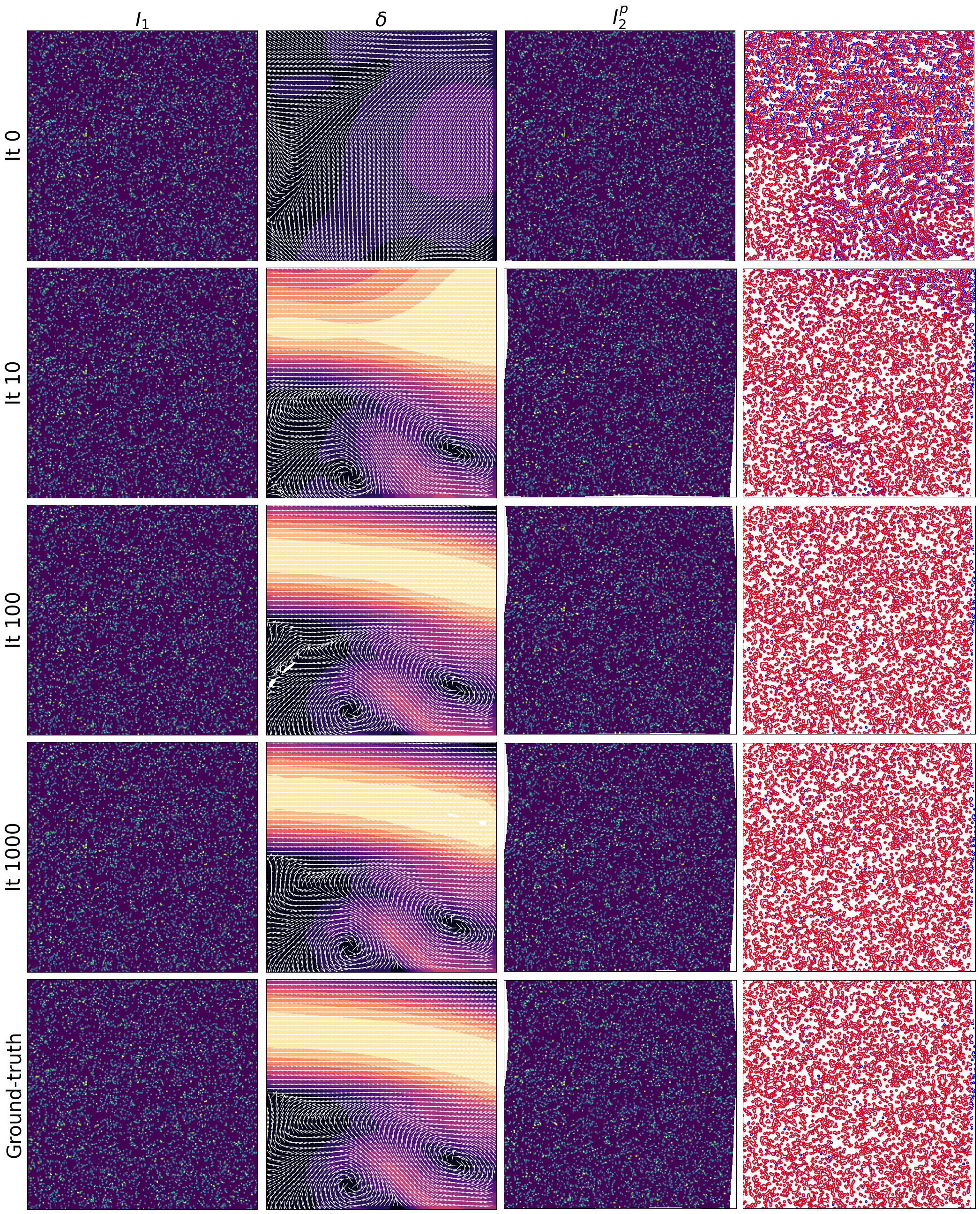}
    \caption{Learning example. From left to right columns, the first image, the inferred displacement, the deformed image, and a superposition of the particles on the deformed image (blue) and the second image (red). Rows are in increasing order of iteration, with the last row showing the ground-truth}
    \label{fig:iterrr}
\end{figure}

Where $N_p$ is the total number of pixels of the image, and $v_i$ are the coordinates for the $i$-pixel. This approach follows the procedure presented by \cite{mingrui2020, Arratia2023}. The ADAM aptimiser\cite{Kingma2017} is used to train the neural network with a learning rate ($\alpha$) defined by the user. A diagram of the iterative nature is presented in \ref{fig:iter}, and an example can be seen in Fig. \ref{fig:iterrr}. Since the number of pixels can make it impossible for $\tilde \delta$ to be trained due to the computational cost, a mini-batch approach is used to train $\tilde \delta$. In this approach, the total number of pixels is randomly separated in batches of size $S_b$, and training is performed over $N$ epochs, with an epoch being defined as the processing of the total number of batches. In Table \ref{tab:conf}, the hyper-parameters and their description are presented.

\begin{table}[H]
    \centering
    \caption{Hyper-parameters}
    \label{tab:conf}
    \begin{tabular}{cl}
    Parameter & Description \\\hline
    \multirow{2}{*}{$\beta$}  & Expected size of the spatial embedding. Allows \\
    &us to control the predominant frequencies.\\
    $N_e$     & Number of spatial embeddings.\\
    $N_l$     & Number of layers.\\
    $S_l$     & Size of the layers.\\
    $\alpha$  & Learning rate.\\
    $S_b$     & Batch size.\\
    $N$     & Epochs.\\
    \end{tabular}%
\end{table}

For demonstration purposes, the results presented in this work were obtained using a Laptop with an NVIDIA GeForce RTX 3070 Laptop GPU (8GB VRAM) and a 12th Gen Intel(R) Core(TM) i7-12700H CPU. Python was selected for the implementation of the proposed methodology, and the PyTorch library was used for the neural networks in the scripts. The code used is made available on \url{https://github.com/emmt1998/NeuralVelocimetry}, where a GUI interface for the proposed methodology is also presented.

\section{Convergence}
Before the validation of the proposed method in flow images is presented, is necessary a small discussion on the convergence of the method. Based on Eq. \ref{eq:loss}, it can be inferred that if the images are mainly formed by constant values, the displacement fields predicted at some coordinates will not be important in the minimisation process of the loss function. As a result, the neural network will converge to a random value at coordinates where no information is present. 
In these cases, the uncertainty of the predictions can be checked by training multiple neural networks with different seeds for the random initialisation of the spatial embedding, the neural network parameters, and the batch permutation. As $\beta$ affects the predominant frequencies on the spatial embedding, it will also influence the smoothness of the results. As $\beta$ is increased, the solutions will take the form of low-frequency displacements. In this case, if there are patches with no information in the images, the values predicted by the neural network will be close to the displacement values where information is present.

On the other hand, if $\beta$ is decreased, the solution will be composed of high-frequency displacements. In these cases, the effects of the patches with no information may worsen, as smaller patches of missing information can be recognized by the embedding. 

\begin{figure}[H]
    \centering
    \includegraphics[width=.8\textwidth]{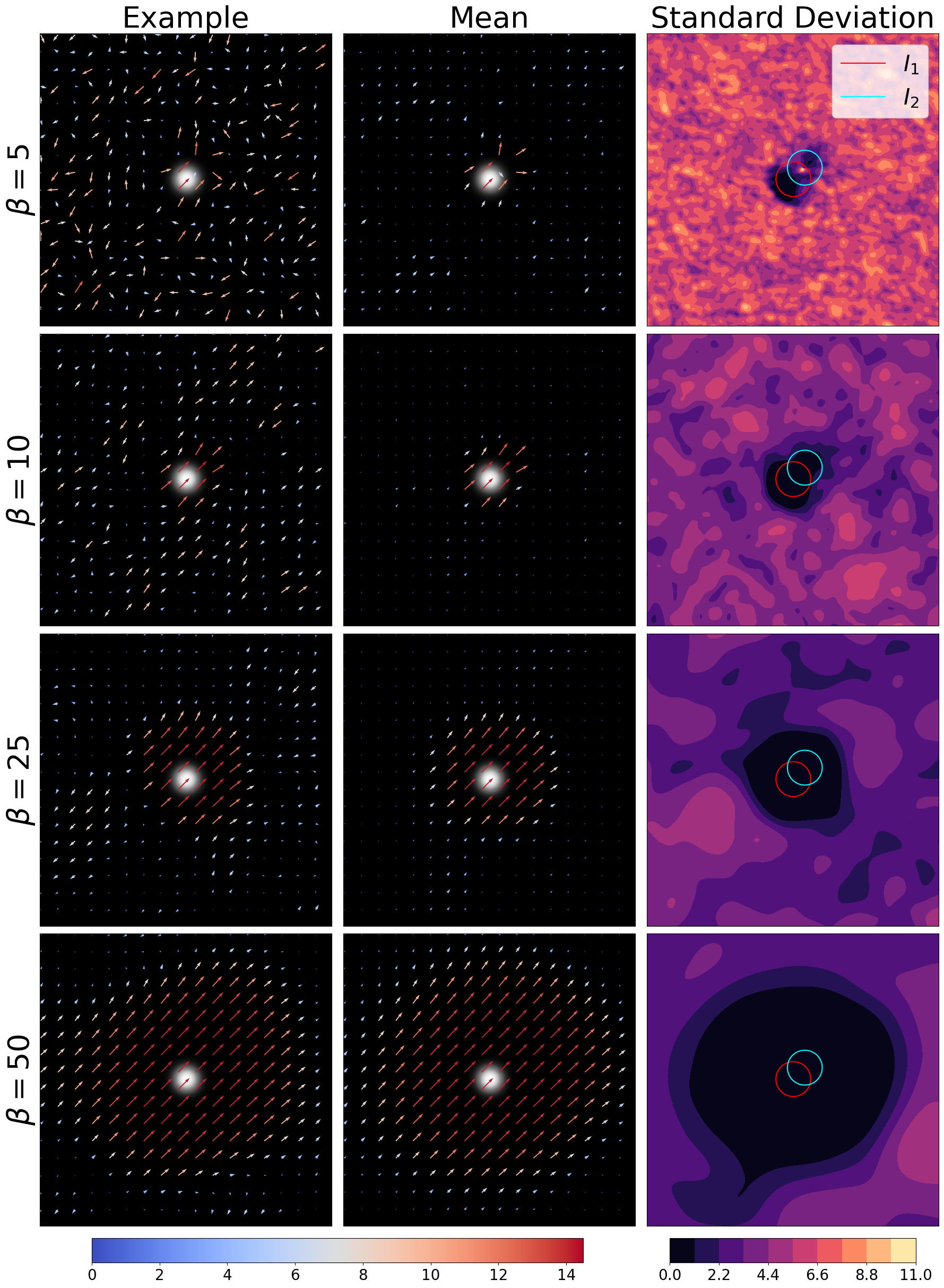}
    \caption{In the middle is the mean displacement field of 10 different initialisations of the method, an example of these results can be seen on the left, and on the right the standard deviation of the results. The effect of $\beta$ can be seen from top to bottom.}
    \label{fig:simpleexample}
\end{figure}

These issues are illustrated by a simple example; a particle in the centre of a $256 \times 256$ px image with a displacement of 10 px in both directions, resulting in a displacement vector of 14.14 px in length. As shown in figure \ref{fig:simpleexample}, as $\beta$ is increased, the results start to show lower frequencies and standard deviation (std), which reduces the overall uncertainty of the neural networks. It can even be seen that if $\beta = 5$, the uncertainty increases where there is an overlap between the particles at both times. This indicates that the frequencies are such that the overlap can be interpreted as a patch with a constant value.

Thus, the results of the proposed methodology will depend on the amount of information available for matching between the images.

\section{Validation on synthetic data}
In this section, the methodology proposed is tested using the synthetic dataset developed by \cite{cai2019} and the synthetic time sequence developed by \cite{Okamoto2000}. The advantage of using synthetic data for testing is that the target displacement fields are known, meaning that the quality of the estimated displacement can be assessed using the Root-Mean-Square Error (RMSE) between the target displacements $\delta$ and the predicted displacements $\tilde \delta$:

\begin{equation}
    \text{RMSE} = \sqrt{\frac{1}{N_p}\sum_{i=1}^{N_p} (\delta_x(\vec{v}_i)-\tilde \delta_x(\vec{v}_i))^2 + (\delta_y(\vec{v}_i)-\tilde \delta_y(\vec{v}_i))^2}
    \label{eq:rmse}
\end{equation}

\subsection{Cai et al. Dataset}
\begin{figure}[H]
    \centering
    \includegraphics[width=\textwidth]{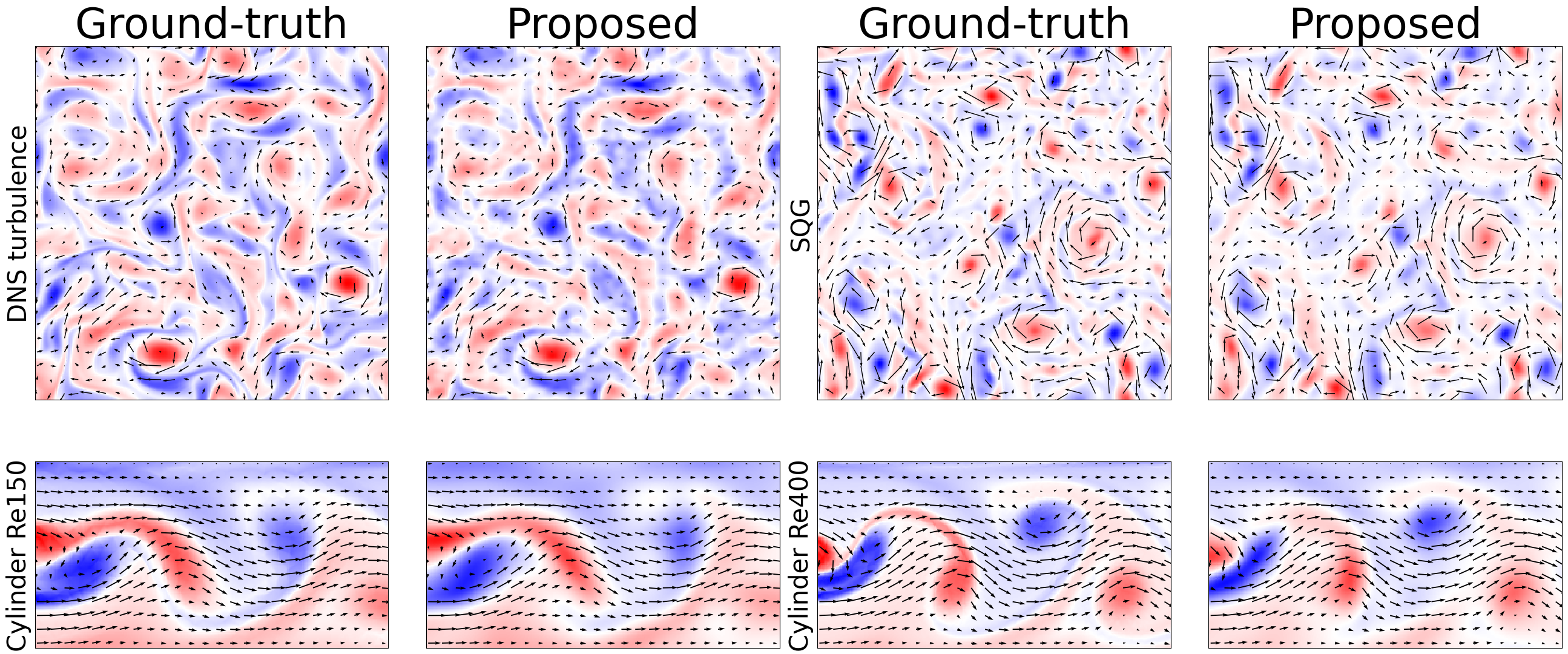}
    \caption{Visual comparison, for illustration purposes between the ground truth (known displacements) and the estimated displacement vectors using the proposed methodology. For better comparison, the vorticity field is shown with the same scale for the colormap for both cases.}
    \label{fig:exacai}
\end{figure}

The Cai et al. Dataset \cite{cai2019} is formed by sequential images with a resolution of $256\times256$ px and their associated known displacement fields. These images were synthetically based on flow fields numerically simulated by  \cite{cai2019} for a uniform flow, a backward-facing step  (Back-step), and the flow after a circular cylinder (Cylinder). Images for different Reynolds number conditions were generated for the Back-step and Cylinder. The dataset also includes images generated from numerical results from other research groups, such as two datasets of a channel flow (forced isotropic turbulence, termed as JHTDB-channel and JHTDB-isotropic1024, respectively), MHD (forced Magnetohydrodynamics turbulence, termed as JHTDB-mhd1024). Also, from simulations of DNS turbulence results (DNS turbulence) and surface quasi-geostrophic (SQG) simulations. It should be highlighted that multiple images were generated for each case, to use on supervised learning approaches. The approach presented in this work learns the displacement fields for a specific pair of images, thus only the first pair of images for each flow case in the dataset is used for the benchmark. The configuration used for each pair differs only in $\beta$, and corresponds to
$N_e=200$,
$N_l=1$,
$S_l=100$,
$\alpha=10^{-3}$,
$S_b=10000$ and 
$N=100$. Learning the displacement vector of each pair of images took approximately 2 seconds. The RMSE obtained for each case, along with the $\beta$ used, is presented in Tab. \ref{tab:otra}. As an example, Fig. \ref{fig:exacai} shows a comparison between the known displacement ('Ground-truth' in the figure) and the displacement obtained using the approach of this work ('Proposed' in the figure).  

\begin{table}[H]
    \centering
    \caption{Summary of the RMSE (pixel) obtained from the synthetic images proposed by Cai et al. 2019. }
    \label{tab:otra}
    \begin{tabular}{lccc}
    
    Case & Cai et al. & Proposed & $\beta$ \\\hline
    Uniform & 0.059 & 0.044 & 200 \\\cline{2-4}
    Back-step $Re=800$ & \multirow{4}{*}{0.072} & 0.072 & \multirow{4}{*}{100}\\
    Back-step $Re=1000$ & &0.099\\
    Back-step $Re=1200$ & &0.057\\
    Back-step $Re=1500$ & &0.074\\\cline{2-4}
    Cylinder $Re=40$ & \multirow{5}{*}{0.115} &0.052&\multirow{5}{*}{50}\\
    Cylinder $Re=150$ & &0.094\\
    Cylinder $Re=200$ & &0.104\\
    Cylinder $Re=300$ & &0.121\\
    Cylinder $Re=400$ & &0.111\\\cline{2-4}
    DNS turbulence & 0.282 &0.112&10\\
    SQG  & 0.294 &0.253&20\\
    JHTDB-channel     & 0.155 &0.087&30\\
    JHTDB-mhd1024    & - &0.150&50\\
    JHTDB-isotropic1024    & - &0.564&20\\
    \end{tabular}%
\end{table}

Based on the results presented in table \ref{tab:otra}, similar results were obtained by the proposed methodology as those obtained by \cite{cai2019}, with slightly better results obtained from the proposed approach in most cases. However, it is necessary to emphasise that the results obtained by \cite{cai2019} were obtained from multiple pairs of images, while the proposed approach is based only on a pair of images. Still, RMSE lower than 1 pixel was produced by the proposed methodology, and this was achieved without prior training on similar flows. It is also necessary to point out that for more turbulent flows, a smaller value of $\beta$ had to be used.

\subsection{Okamoto et al. Jet Shear flow}
For further validation, the Okamoto et al. Jet Shear flow \cite{Okamoto2000} test case is used. Which corresponds to a sequence of 145 synthetic images (frames) with a resolution of 256 $\times$ 256 px will be used, in which the current position of the particles and their true displacement are known. Since the true displacement for each pixel is not known, the RMSE will be calculated by evaluating $\vec{\tilde \delta}$ at the known position of the particles.

As a sequence of images is being used in this case, the neural network will be configured using the first frame until satisfactory results are obtained. The displacements will then be learned sequentially, with the trained neural network parameters of the previous pair being used again as the initial parameters for the neural network of the new pair, speeding up learning and ensuring convergence. By doing this, it is assumed that the displacements in the previous pair of frames should be close to those in the following pair, a fair assumption for sequential data.

\begin{figure}[H]
    \centering
    \includegraphics[width=0.8\textwidth]{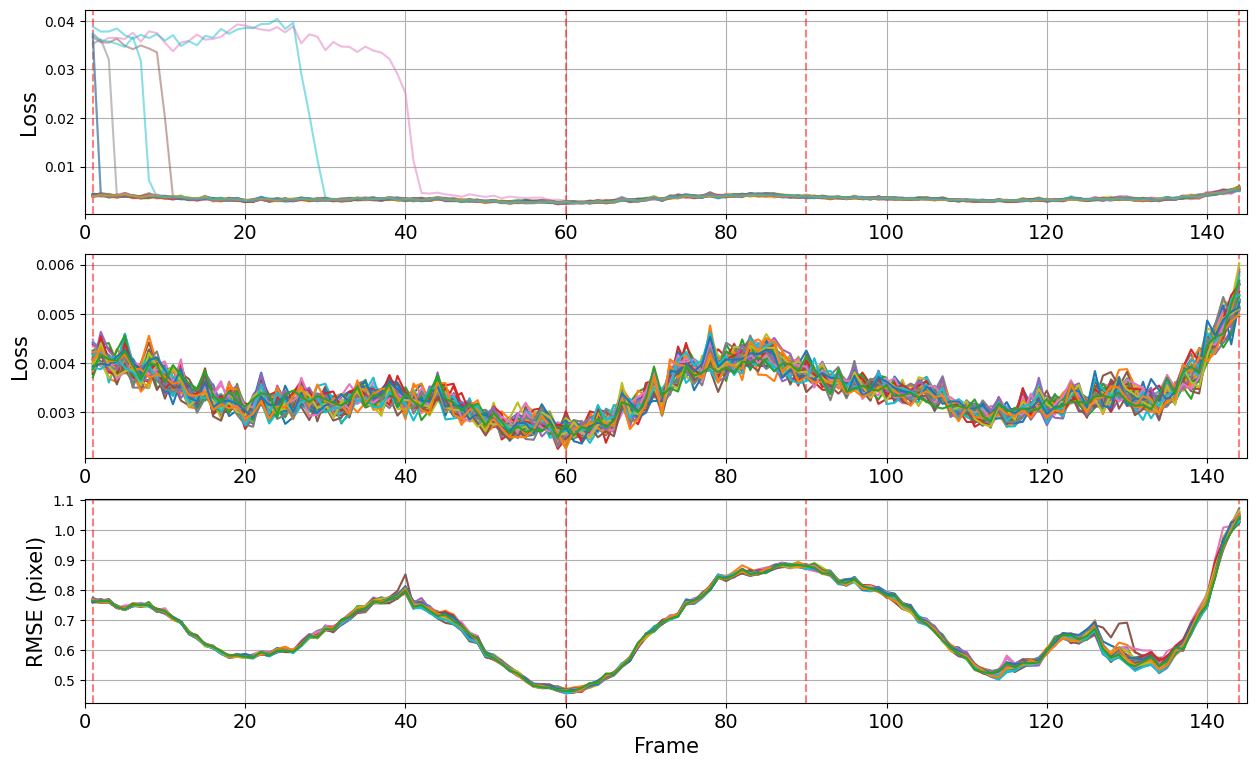}
    \caption{On the top we have the loss for the 50 runs on each frame, in the middle the loss of the 43 runs that converged at the start, in the bottom the RMSE for the displacement of the particles on each frame for the 43 runs that converged. In red, the frames shown on Fig. \ref*{fig:examplejapan}.}
    \label{fig:resultjapan}
\end{figure}

\begin{figure}[H]
    \centering
    \includegraphics[width=\textwidth]{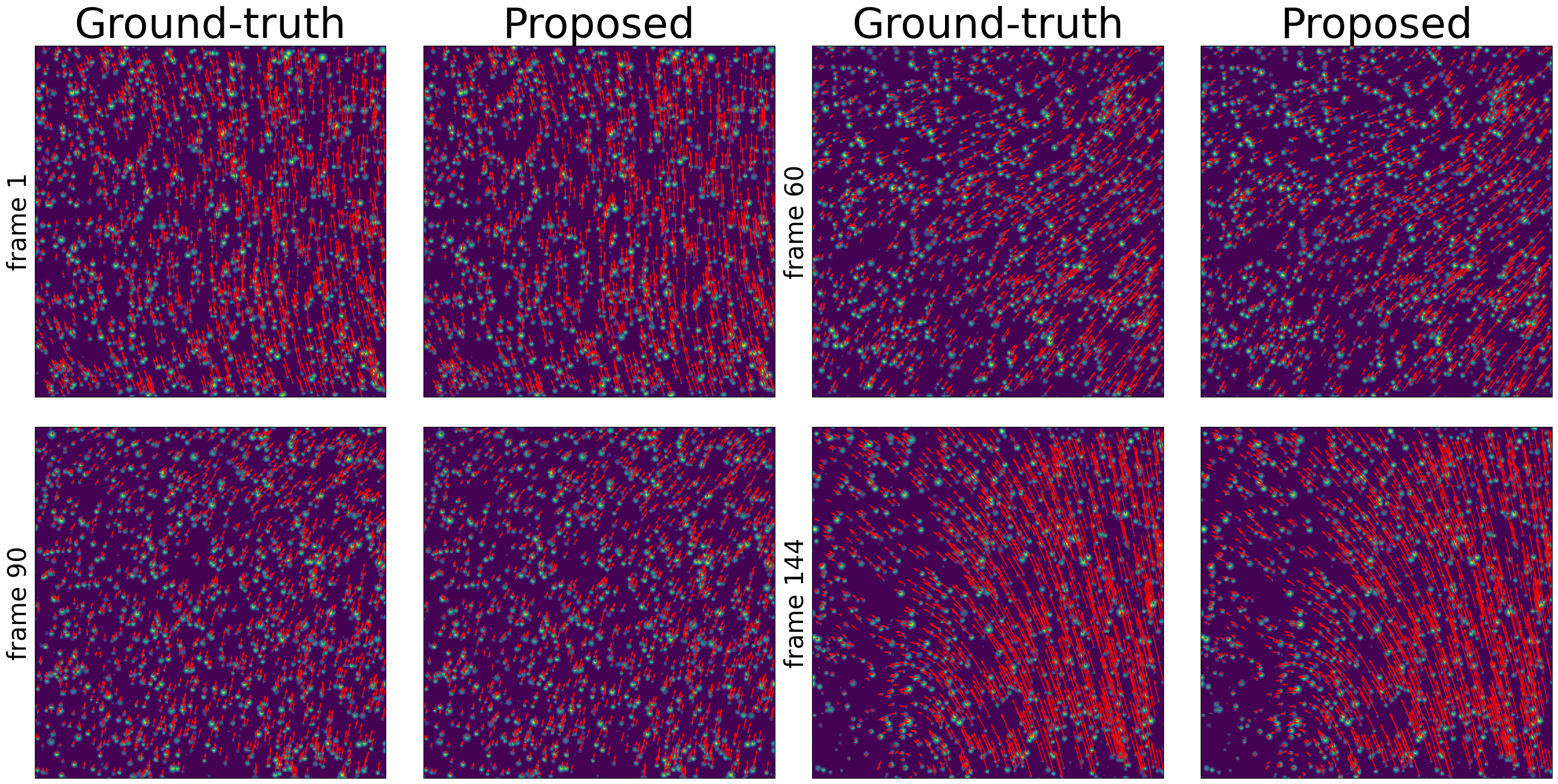}
    \caption{The ground truth and the predicted trajectories for the first run of the 50. In red is the displacement for each particle in the image, both the ground truth and the proposed displacement are on the same pixel scale.}
    \label{fig:examplejapan}
\end{figure}

Additionally, to check the convergence of the method and the possible variation of the RMSE, 50 neural networks will be trained for the full sequence with different random initialisations. This means that, although the neural network is configured to work on the first frame, it could happen that some neural networks do not converge due to specific random initilisations.
The results are shown in Fig. \ref{fig:resultjapan}, with the trajectories on some frames drawn in Fig. \ref{fig:examplejapan}. The configuration used corresponds to 
$\beta=100$,
$N_e=200$,
$N_l=1$,
$S_l=100$,
$\alpha=10^{-3}$,
$S_b=10000$ and 
$N=40$ for the first pair, followed by $N=20$ for subsequent pairs. The 50 runs tooks 64 minutes and 20.8 seconds to train, with an average of 77.2 seconds per run on the entire sequence of 145 frames (144 pairs), and an average of 0.53 seconds per pair of images.

From Fig. \ref{fig:resultjapan}, it can be observed that if the run is continued, it eventually converges, and its loss remains within an accepted range. To filter the runs that did not converge on the first frame, only the loss was used, as a non-converged run shows a loss 10 times greater than that of a converged run. Meanwhile,it is seen that the RMSE remains lower than 1 px, except for the last frames, where it exceeds 1 px. This can be explained by the fact that the displacement in the last frames is greater than in previous intervals, as shown in Fig. \ref{fig:examplejapan}.

It is also noted that all runs achieve similar results for both the loss and the final RMSE, even if they did not converge at the start, with no significant variation observed on any frame for the converged runs. This demonstrates the stability of the results that can be achieved by the proposed methodology. It is important to emphasise the average time required when the warm start approach is used, taking only 0.53 seconds per pair on average. If the parameter of a previous pair are not used to initialise the neural network, an average time 1.06 seconds per pair would have been required, taking approximately 128 minutes for 50 runs on the complete sequence, and without ensuring convergence. 

\section{Application to experimental data}
Lastly, the proposed methodology is tested on a set of experimental data.
\subsection*{Experimental Setup}
\begin{figure}[H]
    \centering
    \includegraphics[height=0.528\textwidth, width=0.8\textwidth]{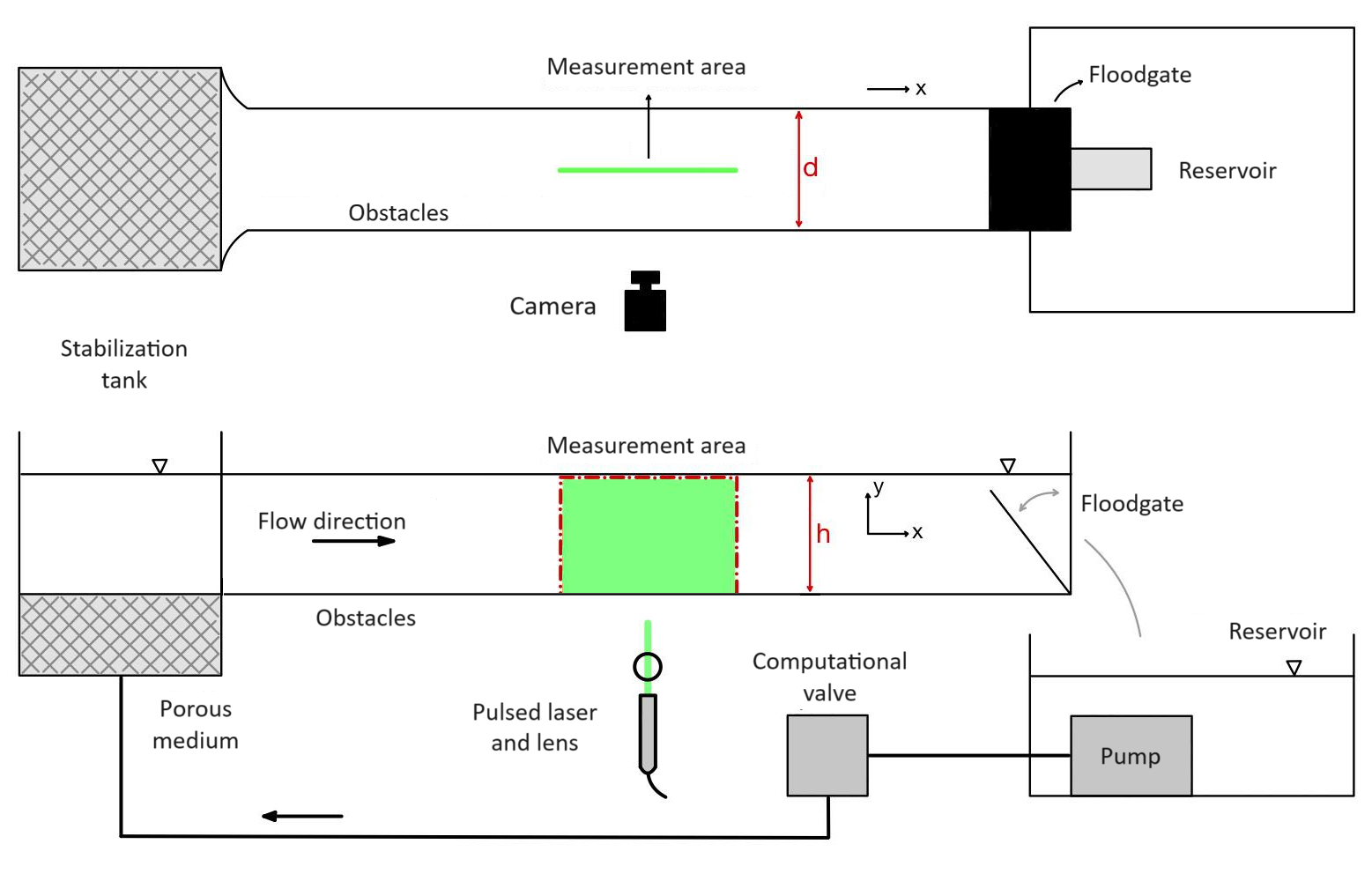}
    \caption{Experimental setup.}
    \label{fig:exp_diagram}
\end{figure}

The experimental case consists of water flowing in an open channel without any obstacles, with a flow rate of 0.033 [m/s], and a Reynolds number of 4660. The channel has a length of 18 [m], a width of 48.6 [cm] and a height of 34 [cm]. A diagram is presented in Fig. \ref{fig:exp_diagram}. Polyamide 12 particles were used as tracers.

A camera resolution of $1501\times2048$ px was used, with a shutter speed of 80 Hz, recording 7200 images (90 seconds). The area of interest was cropped, resulting in a final resolution of $1212\times2048$ px. 
An example of the images taken is shown in Fig. \ref{eximaA}.

\subsection*{Results}
A pre-processing step was applied before the proposed methodology. This step consisted of (in order): the background elimination from the images, the application of a Gaussian filter with a $3\times3$ sized kernel, and the execution of contrast-limited adaptive histogram equalization \cite{Pizer1987}. 

\begin{figure}[H]
    \centering
\begin{subfigure}{0.24\textwidth}
    \centering
    \includegraphics[width=\textwidth]{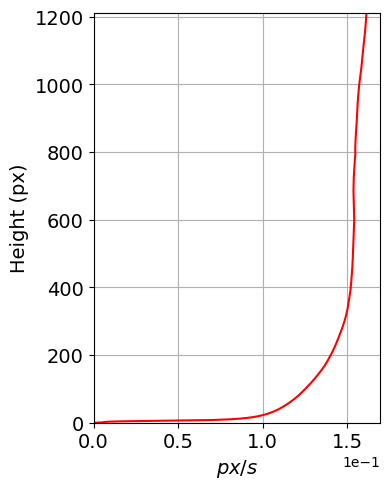}
    \caption{}
\end{subfigure}
\begin{subfigure}{0.24\textwidth}
    \centering
    \includegraphics[width=\textwidth]{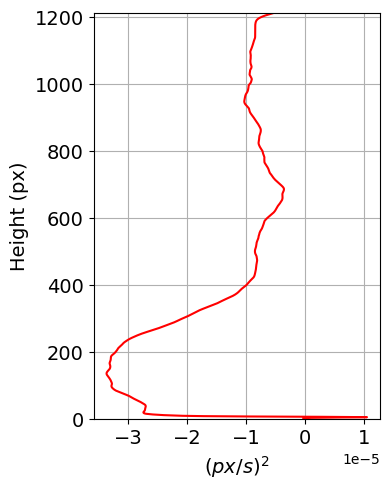}
    \caption{}
\end{subfigure}
\begin{subfigure}{0.24\textwidth}
    \centering
    \includegraphics[width=\textwidth]{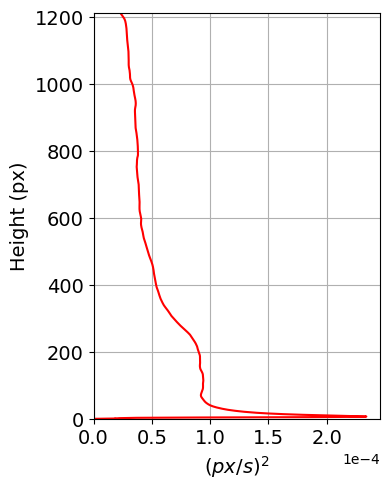}
    \caption{}
\end{subfigure}
\begin{subfigure}{0.24\textwidth}
    \centering
    \includegraphics[width=\textwidth]{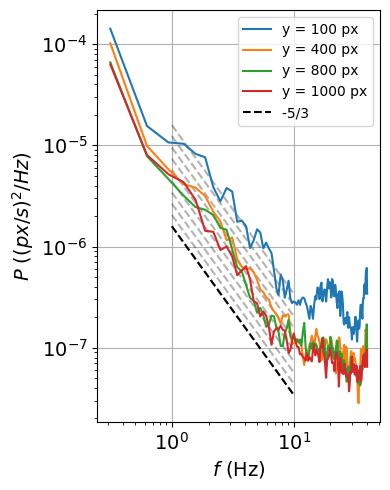}
    \caption{}
    \label{psdfree}
\end{subfigure}
    \caption{(a) Mean Velocity, (b) Reynolds Stress (c) TKE and (d) PSD, calculated at the middle column of the image and in the case of the PSD, choosing the flow at different rows ($y$ on the figure) to perform the spectral analysis.}
    \label{fig:stats2}
\end{figure}

\begin{figure}[h!tbp]
    \centering
\begin{subfigure}{0.45\textwidth}
    \centering
    \includegraphics[width=\textwidth]{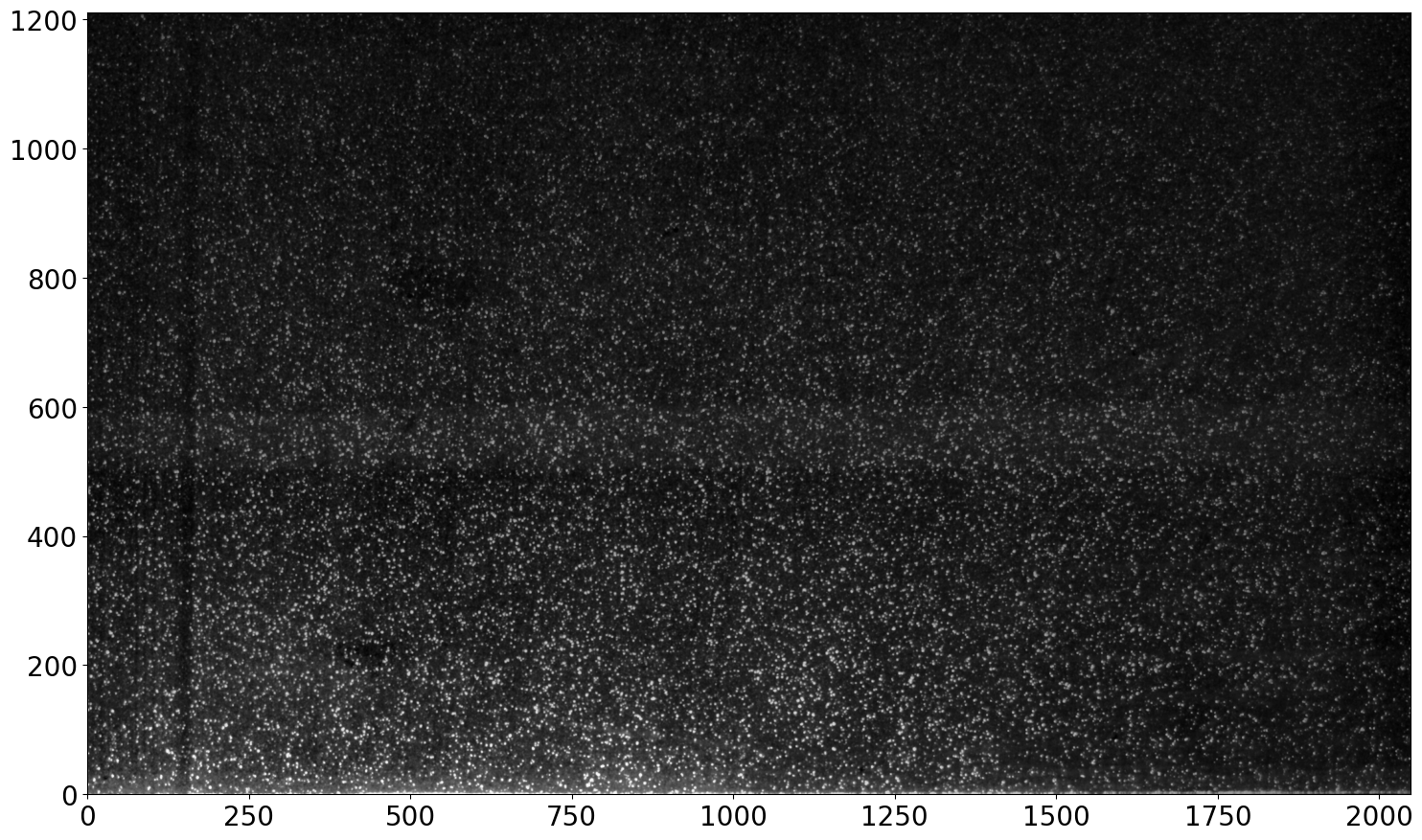}
    \caption{}
    \label{eximaA}
\end{subfigure}
\begin{subfigure}{0.5\textwidth}
    \centering
    \includegraphics[width=\textwidth]{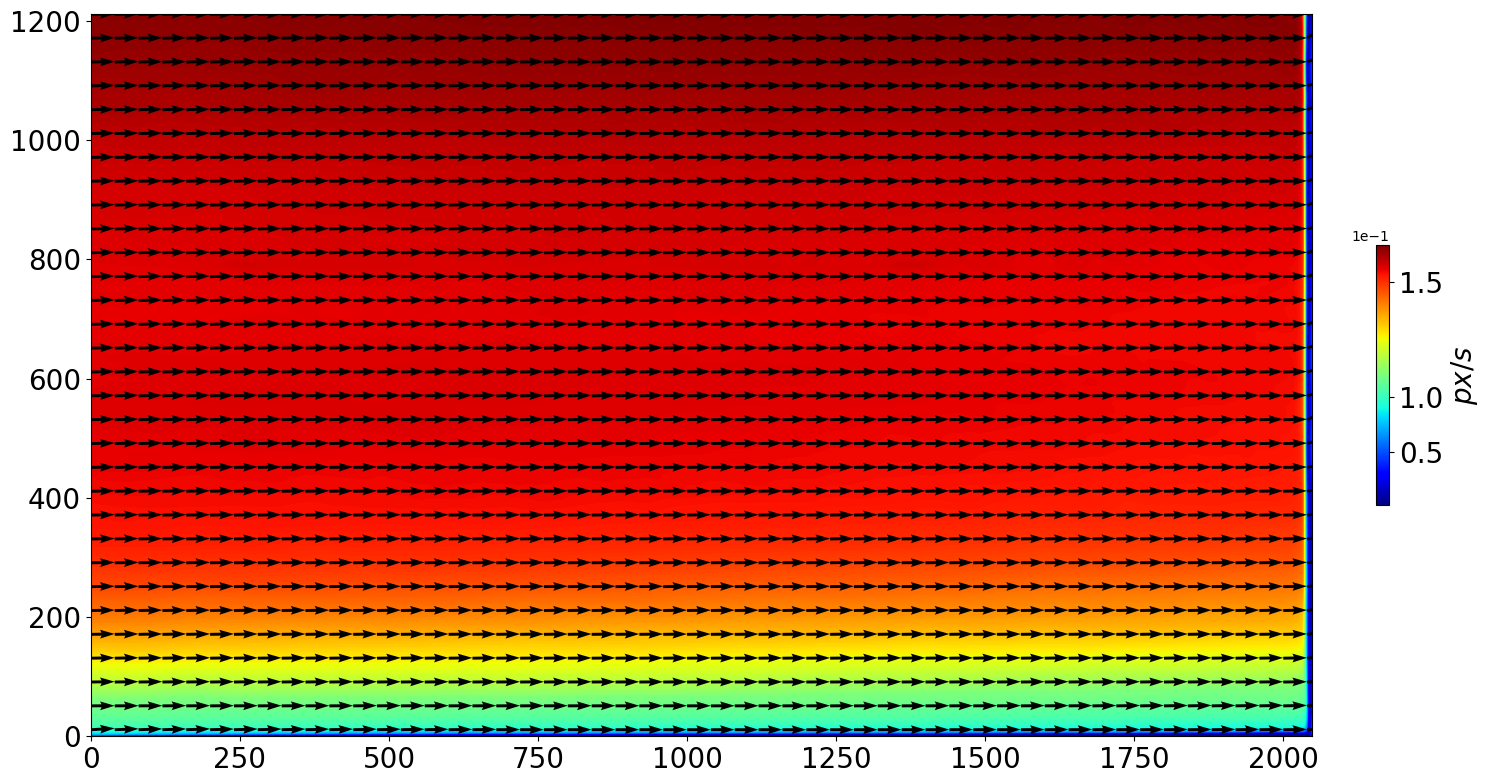}
    \caption{}
    \label{fullmean}
\end{subfigure}

    \caption{(a) Example image and (b) Mean Velocity Field.}
    \label{fig:stats}
\end{figure}

The configuration for the proposed methodology was set to 
$\beta=100$,
$N_e=200$,
$N_l=1$,
$S_l=100$,
$\alpha=10^{-3}$,
$S_b=50000$ and 
$N=10$ for the first pair, followed by $N=1$ for subsequent pairs. The execution time for the full sequence was approximately 180 minutes, with final resolutions of $1212\times2048$ px,
Finally, a statistical analysis of the flow was performed, including the calculation of the mean velocity field, the Reynolds stress, and the Turbulent Kinetic Energy (TKE). These results are shown in Fig. \ref{fig:stats2}, along with the Power Spectral Density (PSD) for different heights. The full mean velocity field is presented in Fig. \ref{fullmean}.

When the time for the full sequence is considered, it results in an average time of 1.5 seconds per pair. An average time 3 times greater than for the Jet Shear flow synthetic case, even though the resolution on the experimental case is 38 greater than the Jet Shear flow case. This difference is attributed to the configuration used in the experimental case, as fewer epochs ($N$) per pair and a greater batch size ($S_b$) were employed.

Looking at the statistical analysis, shown in Fig. \ref{fig:stats2}, the expected behaviour is observed for a free-flow case. The mean velocity is slower near the channel bed, while the TKE increases as the bottom wall is approached. Furthermore, the estimated PSD, presented in Fig. \ref{psdfree}, follows the -5/3 spectral slope predicted by Kolmogorov \cite{Kolmogorov_1962, Lewandowski2016} in a column located in the middle of the image and four vertical positions (rows) of the image.

It should be noted that, although small in size, the proposed methodology produces an outlier zone of vectors on the right of Fig. \ref{fullmean}. This may occur due to the loss function employed, where the first image is compared to the deformed second image. Therefore, if the deformation extends beyond the limits of the image, the first image will compare to zeros in that zone. A potential solution could be the use of a bidirectional loss, meaning the first image would be compared with the deformed second and the second image with the deformed first. However, the effects on the computational cost and time have yet to be studied.

\section{Discussion}
A direct field flow estimation using neural networks and the optical flow equation has been presented. The proposed methodology has been tested in different cases and has shown to be of easy use, as the main adjustments involved modifying $\beta$ and the number of epochs. Good results were achieved in the validation cases, with errors lower or close to 1 px. In the experimental case, the results were found to make physical sense, following the -5/3 spectral slope predicted by Kolmogorov. It is also important to note that the procedure presented returns the displacement field as a continuous function, meaning it can be called for any image coordinates. Therefore, achieving the same resolution as the images used and being able to be evaluated in any image coordinate without the need for any interpolation technique.

However, complications could arise if the resulting vector field is saved with the same resolution as the images. Two channels of 4096 $\times$ 2160 resolution with 32 bits floating-point precision would require 70.77 MB of storage, or approximately 35 MB with some compression algorithm. To avoid this, only $\vec{B}$ and $\vec{\theta}$ could be saved. For the architecture used in all the cases presented, this would require only 203.208 Kb of storage, without compression. In effect, saving the function rather than the vectors.

In particular, the effect of adding a regularisation term to the loss function has not been studied. This could be the focus of future research, as the methodology proposed facilitates the incorporation of regularisation terms into the loss function. It allows for the use of the automatic differentiation capabilities of the neural networks to calculate the derivatives of the flow with respect to x or y, enabling the enforcement of physics knowledge \cite{Raissi2019} which would allow inferring the missing velocity component and/or the pressure with just one neural network \cite{Cai2021, MORENOSOTO2024, DUONG2024, CAI2025}. 

The methodology proposed is not limited to fluids, as the optical flow equation is a generic formulation for any sequence of images. However, further exploration would be required, as the motion of larger rigid solids is better suited to a Lagrangian frame of reference, whereas the methodology presented explains deformation as a field in space, an Eulerian frame of reference.



\section{Acknowledgments}

This work was financially supported by ANID through BECAS/DOCTORADO NACIONAL 21240538.

\bibliographystyle{elsarticle-num} 
\bibliography{citas}
\end{document}